\documentclass{article}

\PassOptionsToPackage{numbers, compress}{natbib}

\usepackage[final]{neurips}

\usepackage{latexsym}
\usepackage{tcolorbox}

\usepackage[T1]{fontenc}
\usepackage[utf8]{inputenc}
\usepackage{microtype}

\usepackage{inconsolata}
\usepackage{wrapfig,lipsum,booktabs}
\usepackage{pifont}
\usepackage{makecell} %multiline cell
\usepackage{tabularx}
\usepackage[normalem]{ulem}
\useunder{\uline}{\ul}{}
\usepackage[export]{adjustbox}

\usepackage{hyperref}       % hyperlinks
\usepackage{url}            % simple URL typesetting
\usepackage{booktabs}       % professional-quality tables
\usepackage{amsfonts}       % blackboard math symbols
\usepackage{nicefrac}       % compact symbols for 1/2, etc.

\usepackage{graphicx}
\usepackage{subfigure}
\usepackage{float}
\usepackage{multirow}

\usepackage{listings}

\usepackage{amsmath}
\usepackage{amssymb}
\usepackage{mathtools}
\usepackage{amsthm}
\usepackage{balance}
\usepackage{flushend}

\usepackage{lipsum}
\usepackage{graphicx}
\usepackage{subcaption}
\usepackage{tikz}
\usetikzlibrary{shapes, arrows, positioning}

\usepackage{fancyhdr}

\hypersetup{
	colorlinks,
	citecolor=gray,
	linkcolor=red,
	urlcolor=blue}

\usepackage[capitalize,noabbrev]{cleveref}
\usepackage{tablefootnote}

\usepackage{color, colortbl,xcolor}
\usepackage{enumitem}
\usepackage{booktabs,arydshln}
\usepackage{xspace}
\usepackage{natbib}
\usepackage{doi}
\usetikzlibrary{arrows.meta, positioning, fit, backgrounds}

\def\paperTitle{\includegraphics[height=14pt]{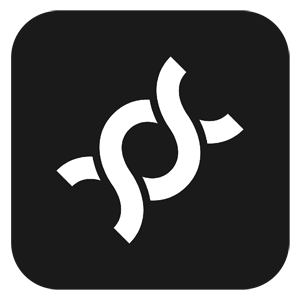} AutoResearch: Insight In, Hallucination Out}

\title{\paperTitle}

\makeatletter
\def\adl@drawiv#1#2#3{%
	\hskip.5\tabcolsep
	\xleaders#3{#2.5\@tempdimb #1{1}#2.5\@tempdimb}%
	#2\z@ plus1fil minus1fil\relax
	\hskip.5\tabcolsep}
\newcommand{\cdashlinelr}[1]{%
	\noalign{\vskip\aboverulesep
		\global\let\@dashdrawstore\adl@draw
		\global\let\adl@draw\adl@drawiv}
	\cdashline{#1}
	\noalign{\global\let\adl@draw\@dashdrawstore
		\vskip\belowrulesep}}
\makeatother

\setlist[itemize]{align=parleft,left=0pt..0.5em}
\setlist[enumerate]{align=parleft,left=0pt..0.5em}

\setlist[itemize]{align=parleft,left=0pt..0.8em}

\usepackage{booktabs}
\newcommand{\evomapheader}{%
  \noindent
  \begin{minipage}[c]{0.62\linewidth}
    \href{https://evomap.ai}{%
      \includegraphics[
        height=14pt,
        keepaspectratio
      ]{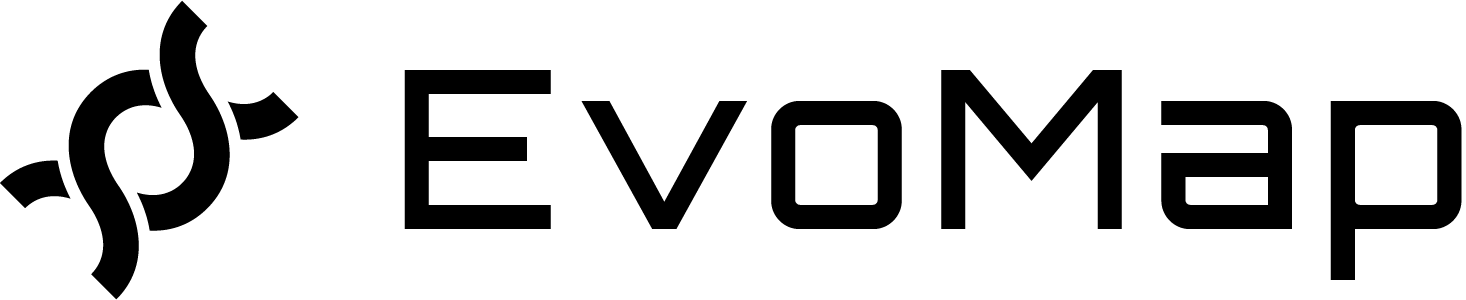}}
  \end{minipage}%
  \hfill
  \vspace{4pt}
}

\def\authorBlock{
    Xiang Liu \qquad Qumeng Sun \qquad  Bruno \qquad Xiao Zhang  \qquad Jiahao Li\\
    Project Leader: \quad \textbf{Haoyang Zhang}\footnotemark[2]\\
    Infinite Evolution Lab, EvoMap\\
    \includegraphics[height=10pt]{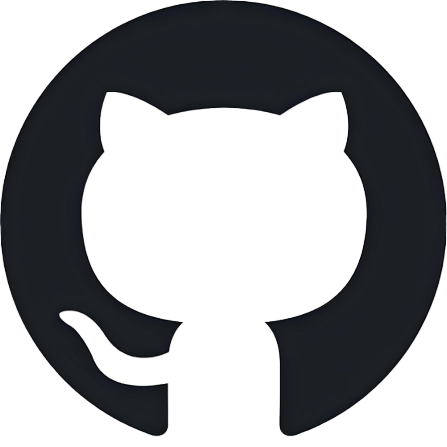} \url{https://github.com/EvoMap/AutoResearch}
}

\author{\authorBlock}

\begin{document}
\evomapheader
\vspace{-0.08in}
\maketitle

{
  \renewcommand{\thefootnote}%
  {\fnsymbol{footnote}}
  \footnotetext[2]{Corresponding Author. {\tt\small 17@evomap.ai}}
}

\begin{abstract}
Autonomous research systems are increasingly capable of executing long research workflows, yet automation alone does not ensure that the resulting process remains scientifically grounded.
We introduce \textbf{AutoResearch}, a two-stage system that connects \textit{Idea Generation} with \textit{Idea Execution} to address both how research ideas are formed and how they are reliably established through experimentation.
In Idea Generation, AutoResearch continuously integrates emerging research signals with accumulated domain knowledge, identifies transferable mechanistic insights, and uses multi-model generation and cross-review to produce grounded, testable research plans.
In Idea Execution, coordinated agents decompose these plans into experiments, iteratively implement and diagnose them, and employ independent evidence-based review before accepting research conclusions.
Across representative settings in cross-modal retrieval, systems optimization, and benchmark-driven machine learning, AutoResearch turns generated ideas into measurable progress, detects and corrects unreliable experimental results, and makes evidence-conditioned decisions to continue, revise, or terminate research directions.
For example, on RSICD benchmark, an AutoResearch-generated idea improves mean Recall from $32.84$ to $34.69$, while recording only $5$ audit-confirmed issue events compared with $11$--$27$ for other autonomous research systems.
These results demonstrate a research process in which meaningful insight is grounded before experimentation and conclusions are grounded before acceptance: \textit{Insight In, Hallucination Out}.
\end{abstract}

\section{Introduction}
\label{sec:intro}

Recent advances in large language models and agentic systems have substantially expanded the range of scientific research tasks that can be delegated to AI~\citep{lu2024aiscientist,schmidgall2025agentlaboratory}.
Recent autonomous research systems can further carry increasingly long research workflows from high-level objectives to hypothesis formulation, implementation, experimentation, and research artifacts~\citep{yamada2025aiscientistv2,tang2025airesearcher,mitchener2025kosmos}.
Taken together, much of this progress has advanced along an \textit{automation} dimension: extending the portion of the research workflow that can be delegated to agents and reducing the human intervention required during execution~\citep{mitchener2025kosmos,yang2025rdagent}.
However, automating a larger fraction of the workflow does not by itself ensure a scientifically meaningful research process.
A complementary \textit{research} dimension concerns whether the system can continuously build and update research knowledge, transform that knowledge into research directions worth pursuing, and reliably develop those directions into evidence-supported outcomes.
From this perspective, autonomous research requires an integrated system spanning \textit{research discovery} and \textit{research execution}, where automation serves rather than defines the process.

Importantly, spanning \textit{research discovery} and \textit{research execution} requires more than simply connecting the two stages, as they impose distinct grounding requirements on an autonomous research system.
In research discovery, the system must maintain an evolving research context, identify technically meaningful signals, and determine whether mechanisms emerging elsewhere can motivate new directions in the target domain.
Existing systems often operate from a user-provided idea, objective, task, or predefined experimental context~\citep{lu2024aiscientist,schmidgall2025agentlaboratory,yang2025rdagent,liu2026autoresearchclaw}, while recent efforts have extended toward autonomous hypothesis generation and open-ended exploration~\citep{yamada2025aiscientistv2,tang2025airesearcher,mitchener2025kosmos}.
However, continuously accumulating research knowledge and systematically transforming emerging signals into domain-grounded hypotheses remain comparatively underexplored~\citep{tang2026narrow}, making the preservation of mechanistic \textit{insight} a central challenge.
In research execution, an idea must be translated into code, iteratively refined, experimentally tested, and ultimately supported by evidence consistent with the intended hypothesis.
Although recent systems have introduced iterative research--development loops, failure recovery, and explicit verification~\citep{yang2025rdagent,liu2026autoresearchclaw}, errors in implementation, measurement, or interpretation can still propagate into coherent but unsupported research claims~\citep{trehan2026llmscientists,ding2026verificationgap}, which we view as a system-level form of \textit{hallucination}.
These requirements lead to the central questions of this work: \textbf{how can an autonomous system continuously derive meaningful research insight from an evolving knowledge landscape, and how can it prevent hallucination as that insight is transformed into experimental conclusions?}

To address these requirements, we develop \textbf{AutoResearch}, a two-stage research system built around \textit{knowledge-grounded research discovery} and \textit{evidence-grounded research execution}.
Rather than treating idea generation and experimentation as isolated capabilities, AutoResearch connects them into a unified process that transforms evolving research knowledge into grounded ideas and grounded ideas into evidence-backed outcomes.
The discovery stage combines external research signals with accumulated domain knowledge to identify transferable mechanisms and formulate testable ideas, while the execution stage carries selected ideas through planning, implementation, experimentation, diagnosis, and verification.
Conceptually, AutoResearch follows a compact trajectory of \textit{research signals + knowledge $\rightarrow$ grounded ideas $\rightarrow$ evidence-backed outcomes}.
This design gives concrete meaning to \textit{Insight In, Hallucination Out}: insight is grounded before it motivates research, and conclusions are grounded before they are accepted as research outcomes.

AutoResearch operationalizes both stages through coordinated multi-agent workflows, allowing exploration, execution, and verification to be distributed across specialized agents rather than compressed into a single model trajectory.
In the discovery stage, heterogeneous research signals are progressively filtered and grounded against a curated knowledge base, after which multiple frontier models independently generate cross-domain hypotheses, review one another's proposals, assess freshness and domain consistency, and convert surviving ideas into executable research plans.
This design turns broad information acquisition into a selective process in which increasingly strong evidence is required before a signal becomes a research commitment.
In the execution stage, planning, implementation, experimentation, diagnosis, review, and verification are similarly decomposed across agents that operate over shared and persistent research state.
When broader exploration or parallel execution is beneficial, these workflows can further expand into swarm-style coordination, where multiple agents jointly explore alternatives, claim tasks, exchange intermediate results, and independently verify critical outcomes.
Together, these mechanisms allow AutoResearch to scale both the search for promising ideas and the effort required to establish them, while preserving a traceable path from research intent to experimental evidence.

We evaluate AutoResearch across three representative scenarios commonly encountered in research practice: open-ended method exploration, systems optimization with measurable objectives, and benchmark-driven machine learning.
In cross-modal retrieval, an autonomously generated and cross-reviewed research idea translates into consistent staged improvements from $32.84$ to $34.69$ mR ($+1.85$), demonstrating that discovered insights can be concretely instantiated and experimentally supported.
Across the same evaluation, AutoResearch records only $5$ issue events, compared with $11$--$27$ for the other autonomous research systems, while the remaining studies further demonstrate its ability to diagnose unreliable measurements and make evidence-conditioned decisions to continue, revise, or terminate an experiment.
Together, these results illustrate how AutoResearch connects research discovery with reliable execution, turning emerging knowledge into experimentally grounded progress while preventing unsupported claims from propagating through the research process.

\section{AutoResearch}

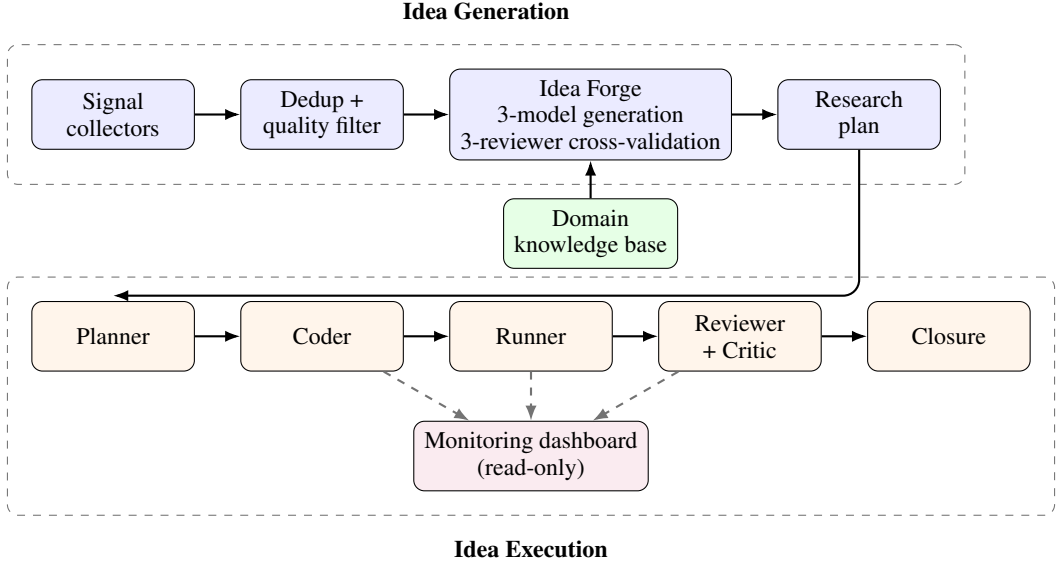
\begin{figure}[t]
  \centering
  \begin{tikzpicture}[
      node distance=0.7cm and 0.6cm,
      box/.style={
        draw,
        rounded corners,
        minimum height=0.92cm,
        minimum width=2.15cm,
        align=center,
        font=\small,
        inner sep=4pt
      },
      stage/.style={
        draw=black!55,
        dashed,
        rounded corners,
        inner sep=9pt
      },
      arr/.style={-{Latex[length=2mm]}, thick},
      mon/.style={-{Latex[length=2mm]}, thick, dashed, draw=black!55}
    ]

    % -------------------------
    % Stage 1: Idea Generation
    % -------------------------
    \node[box, fill=blue!8] (collect) at (0,0) {Signal\\collectors};
    \node[box, fill=blue!8, right=of collect] (filter) {Dedup +\\quality filter};
    \node[box, fill=blue!8, right=of filter, minimum width=3.1cm] (forge)
      {Idea Forge\\3-model generation\\3-reviewer cross-validation};
    \node[box, fill=blue!8, right=of forge] (plan0) {Research\\plan};

    \node[stage, fit=(collect)(filter)(forge)(plan0),
      label={[font=\small\bfseries,yshift=2mm]above:Idea Generation}] (s1) {};

    % Shared knowledge base
    \node[box, fill=green!10, below=0.5cm of forge] (kb) {Domain\\knowledge base};
    \draw[arr] (kb) -- (forge);

    % -------------------------
    % Stage 2: Idea Execution
    % -------------------------
    \node[box, fill=orange!8, below=2.0cm of collect] (plan1) {Planner};
    \node[box, fill=orange!8, right=of plan1] (code) {Coder};
    \node[box, fill=orange!8, right=of code] (run) {Runner};
    \node[box, fill=orange!8, right=of run] (review) {Reviewer\\+ Critic};
    \node[box, fill=orange!8, right=of review] (close) {Closure};

    % Monitoring dashboard
    \node[box, fill=purple!8, below=0.65cm of run, minimum width=2.8cm] (dash)
      {Monitoring dashboard\\(read-only)};

    \node[stage, fit=(plan1)(code)(run)(review)(close)(dash),
      label={[font=\small\bfseries,yshift=-2mm]below:Idea Execution}] (s2) {};

    % -------------------------
    % Arrows
    % -------------------------
    \draw[arr] (collect) -- (filter);
    \draw[arr] (filter) -- (forge);
    \draw[arr] (forge) -- (plan0);

    % plan handoff to execution
    \draw[arr, rounded corners=6pt]
      (plan0.south) |- ([yshift=2pt]plan1.north);

    \draw[arr] (plan1) -- (code);
    \draw[arr] (code) -- (run);
    \draw[arr] (run) -- (review);
    \draw[arr] (review) -- (close);

    % monitoring links
    \draw[mon] (code) -- (dash);
    \draw[mon] (run) -- (dash);
    \draw[mon] (review) -- (dash);

  \end{tikzpicture}
  \caption{AutoResearch consists of two stages: idea generation converts external signals and domain knowledge into a research plan, and idea execution develops the plan through a resumable multi-agent workflow. The monitoring dashboard is a read-only layer over execution states.}
  \label{fig:architecture}
\end{figure}

As shown in~\cref{fig:architecture}, we formulate \textbf{AutoResearch} as a two-stage research process consisting of \textit{Idea Generation} and \textit{Idea Execution}.
The first stage transforms an evolving research context into a grounded and testable research idea, while the second stage develops that idea into an outcome supported by explicit experimental evidence.

\subsection{Problem Formulation}

Let $\mathcal{D}$ denote a target research domain, and let the research context available at time $t$ be
\begin{equation}
\mathcal{C}_t = \left(\mathcal{K}_t, \mathcal{S}_t\right),
\end{equation}
where $\mathcal{K}_t$ represents accumulated domain knowledge and $\mathcal{S}_t$ represents newly observed signals from the evolving research landscape.

\textit{Idea Generation} seeks a mapping
\begin{equation}
G_{\mathrm{idea}}: \left(\mathcal{C}_t, \mathcal{D}\right) \rightarrow (h, p),
\end{equation}
where $h$ is a research hypothesis and $p$ is an executable research plan.
We consider an idea grounded when its motivation is traceable to the available research context, its connection to the target domain is supported by a meaningful mechanism rather than superficial association, and its central hypothesis can be empirically tested.
This grounding requirement defines the role of \textit{insight}: the system should identify not only what appears novel, but why a research direction is worth pursuing.

\textit{Idea Execution} is formulated as a stateful process initialized from $(h,p)$.
At execution step $k$, the system performs a research action $a_k$, observes the resulting environment output or artifact $o_k$, and updates its research state as
\begin{equation}
x_{k+1} = F(x_k, a_k, o_k),
\qquad
x_0 = (h,p).
\end{equation}

The state $x_k$ summarizes the evolving plan, implementation, experimental observations, critiques, evidence, and research decisions accumulated during execution.
Let $\mathcal{E}$ denote the resulting evidence set and $c$ a research claim produced from this process.
A claim is accepted only when it is supported by corresponding execution evidence,
\begin{equation}
\operatorname{Accept}(c)
\Rightarrow
\exists e \in \mathcal{E}
; \operatorname{Supports}(e,c)=1.
\end{equation}

This evidence-grounding requirement captures the operational meaning of \textit{Hallucination Out}: unsupported intermediate outputs should not be promoted into established research conclusions.

The two stages therefore impose complementary grounding requirements.
\textit{Idea Generation} asks why a hypothesis deserves to be tested, while \textit{Idea Execution} asks why an experimental conclusion deserves to be accepted.
The following sections describe how AutoResearch operationalizes these two requirements through coordinated multi-agent research workflows.

\subsection{How Does AutoResearch Generate a Research Idea?}

As illustrated in~\cref{fig:architecture}, the Idea Generation stage transforms heterogeneous research signals and accumulated domain knowledge into a grounded hypothesis and an executable research plan.
AutoResearch grounds this process in two complementary sources of context: an evolving stream of external research signals and a curated domain knowledge base.
The former captures what the broader research community is newly observing, discussing, and developing, while the latter represents what is already known within the target domain.
Rather than generating novelty directly from either source, AutoResearch seeks technically meaningful mechanisms that can be transferred across research contexts and formulated into testable hypotheses.

\subsubsection{Building an Evolving Research Context}

AutoResearch maintains two complementary sources of context: newly emerging research signals and accumulated domain knowledge.
External signals are continuously collected from research communities, papers, repositories, and technical media, including fast-moving social platforms such as X and Xiaohongshu.
Importantly, AutoResearch does not treat all sources as equally informative.
Researchers, practitioners, and technical creators with consistently strong curatorial judgment often surface promising methods, engineering observations, and cross-domain connections earlier than conventional literature search, providing a useful source-level prior for downstream idea discovery.
The system therefore combines source-aware quality priors with normalization, deduplication, and model-based screening to retain signals with substantive technical content.
In parallel, a curated domain knowledge base summarizes what is already known in the target research area.
Together, the two sources capture both \textit{what the field already knows} and \textit{what high-quality observers are newly noticing}, providing a continually updated context for Idea Forge.

\subsubsection{From Research Signals to Transferable Insights}

A research seed is useful only when it contains more than a topic or a reported improvement.
AutoResearch instead seeks an identifiable \textit{mechanistic insight}, namely a method, observation, or relation whose technical rationale can be separated from its original setting and examined in another research context.
This criterion biases the system toward signals that suggest \textit{why} something works, rather than signals that merely report that it works.

Given a selected seed $s$ and a target domain $d$, AutoResearch formulates idea generation as a mechanism-transfer problem,
\begin{equation}
\mathcal{H}(s,d)=\left\{G_m!\left(s,\mathcal{K}^{(d)}\right)\right\}_{m=1}^{M},
\end{equation}
where $G_m$ denotes an independent generator and $\mathcal{H}(s,d)$ is the resulting set of candidate hypotheses.
Each generator is asked to determine whether the mechanism represented by $s$ can address an unresolved problem in $d$, rather than simply combining terminology from the two contexts.
A generator may reject the pairing when no substantive transfer can be identified.
This explicit \textit{no-match} outcome prevents every observed signal from being forced into a research proposal.

\subsubsection{From an Insight to an Executable Research Idea}

Candidate hypotheses are subjected to independent cross-validation before they become research commitments.
In the current implementation, three frontier models independently generate proposals and three reviewers assess each candidate, with at least two positive reviews required for the idea to advance.
The review focuses on whether the transferred mechanism is technically meaningful, the proposed method remains sufficiently simple to isolate its effect, and the hypothesis can be tested under a realistic experimental protocol.

Surviving ideas are subsequently checked against recent research and the target-domain knowledge base. 
A freshness step updates stale models, benchmarks, or references when necessary, while a domain-consistency check ensures substantive use of target-domain knowledge. 
Multi-model generation and independent review provide diversity and cross-checking, while the decomposition across signals, domains, and hypotheses naturally supports parallel multi-agent or swarm-style exploration. 
Finally, AutoResearch converts each validated hypothesis into an executable research plan whose initial experiment determines whether the direction merits further investment. 
The output of Idea Generation is therefore a pair $(h,p)$ consisting of a grounded hypothesis and a concrete plan for testing it.

\subsection{How Does AutoResearch Execute a Research Idea?}

As illustrated in~\cref{fig:architecture}, Idea Execution takes a validated hypothesis $h$ and research plan $p$ and develops them through a stateful, evidence-grounded workflow.
The central design principle is to separate \textit{producing} a research result from \textit{establishing} that the result is valid.
AutoResearch therefore decomposes a research plan into executable units, iteratively updates them with observations from the real environment, and subjects critical outcomes to independent review before they are accepted as research conclusions.

\subsubsection{From a Research Plan to Decomposable Execution}

Research experiments are often naturally decomposable into implementation, pilot evaluation, ablation, diagnosis, and validation tasks with explicit dependencies.
We represent a research plan as a task graph
\begin{equation}
\mathcal{G}_{p} = (\mathcal{T}, \mathcal{R}),
\end{equation}
where $\mathcal{T}={\tau_1,\ldots,\tau_n}$ denotes executable research tasks and $\mathcal{R}$ specifies their dependency relations.
At execution step $k$, the coordinator identifies tasks whose prerequisites have been satisfied,
\begin{equation}
\mathcal{T}^{\mathrm{ready}}_k
=
\left\{
\tau_i \in \mathcal{T}
\mid
\mathrm{Pred}(\tau_i)
\subseteq
\mathcal{T}^{\mathrm{done}}_k
\right\},
\end{equation}
and dispatches independent tasks concurrently when their dependencies permit.

Each executed action $a_k$ produces an observation or artifact $o_k$ from the actual environment and updates the persistent research state,
\begin{equation}
x_{k+1}=F(x_k,a_k,o_k).
\end{equation}

The state records the evolving plan, implementation, experimental results, reviews, and decisions, allowing a project to resume from verified progress rather than from an opaque conversation history.
This decomposition also provides a natural basis for multi-agent execution and, when broader parallelism is useful, swarm-style coordination.

\subsubsection{How Are Errors Detected and Corrections Triggered?}

AutoResearch deliberately separates implementation from critical evaluation.
The reviewer examines whether the implementation and experimental protocol remain faithful to the original hypothesis, while the critic assesses whether the observed evidence is sufficient to support the emerging research claim.
Crucially, critical evaluation is performed by a \textit{fresh-context} agent that does not inherit the producer's reasoning trajectory.
Instead, it receives only the information required for independent assessment, such as the hypothesis, research plan, experimental artifacts, and evaluation criteria.
This design reduces anchoring to earlier decisions and forces the evaluator to reconstruct the validity of the result from the research record itself.

Let $\mathcal{E}_k$ denote the evidence accumulated by step $k$ and $\Gamma$ the corresponding evaluation criteria.
The independent evaluator produces
\begin{equation}
v_k
=
V(h,p,\mathcal{E}_{k},\Gamma)
\in
{\mathsf{PASS},\mathsf{PARTIAL},\mathsf{FAIL}}.
\end{equation}

A semantic mismatch, failed criterion, or insufficiently supported result triggers a correction path rather than being absorbed into the subsequent narrative,
\begin{equation}
v_k \neq \mathsf{PASS}
\quad\Longrightarrow\quad
a*{k+1}
\in
{\mathsf{DIAGNOSE},\mathsf{REVISE},\mathsf{RERUN}}.
\end{equation}

The workflow therefore treats unexpected and negative observations as signals for diagnosis and decision making rather than as obstacles to be explained away.

\subsubsection{When Does Evidence Support a Research Claim?}

AutoResearch distinguishes an experimental output from admissible research evidence.
For an executed action $a_i$, we represent an evidence item as
\begin{equation}
e_i=(a_i,o_i,r_i),
\end{equation}
where $o_i$ is the observed environment output and $r_i$ denotes the corresponding persistent artifact, such as an evaluation record, log, checkpoint, or result file.
For a research claim $c$, the system identifies the relevant evidence subset $\mathcal{E}_c \subseteq \mathcal{E}$ and requires independent verification before the claim is accepted,
\begin{equation}
\operatorname{Accept}(c)
\Rightarrow
V(c,\mathcal{E}_c,\Gamma_c)=\mathsf{PASS}.
\end{equation}

In this sense, execution produces observations, while verification determines whether those observations constitute sufficient evidence for a claim.

Evidence also determines how the research process proceeds.
Depending on the current result, AutoResearch may \textsc{Continue}, \textsc{Revise}, \textsc{Scale}, or \textsc{Stop}; termination may record a supported, falsified, or inconclusive hypothesis.
A negative result is therefore a valid research outcome when it is supported by evidence, whereas an unsupported positive result is not.
This separation between execution, verification, and research closure provides the operational basis for \textit{Hallucination Out}: plausible intermediate outputs cannot become research conclusions solely because the agents that produced them consider the task successful.

The execution states, reviews, and evidence records are additionally exposed through the read-only monitoring dashboard in~\cref{fig:architecture}, providing visibility into the process without participating in research decisions.

\section{Evaluating AutoResearch as a Research System}
\label{sec:cases}

\subsection{Evaluation Setup and Comparison Systems}

We evaluate AutoResearch in three representative scenarios commonly encountered in research practice: \textit{open-ended method exploration}, \textit{systems optimization with explicit measurable constraints}, and \textit{benchmark-driven machine learning}.
These settings respectively examine whether the system can develop a research idea into measurable progress, detect and correct unreliable experimental outcomes, and make evidence-conditioned decisions about whether a research direction should continue, be revised, or terminate.

For the comparative settings, we consider four representative autonomous research systems: The AI Scientist~\citep{lu2024aiscientist}, Agent Laboratory~\citep{schmidgall2025agentlaboratory}, R\&D-Agent~\citep{yang2025rdagent}, and AutoResearchClaw~\citep{liu2026autoresearchclaw}.
Together, they cover several major paradigms of autonomous research, including end-to-end scientific workflows, execution from human-provided research ideas, iterative research--development cycles driven by experimental feedback, and recent multi-agent systems with failure recovery and result verification.
For each comparative scenario, all systems receive the same research objective and are evaluated under the same task-specific experimental contract.
We assess not only the reported task outcome, but also the reliability of the process by auditing the produced research artifacts and recording confirmed issue events, thereby avoiding reliance on each system's self-reported conclusion alone.
The benchmark-driven machine-learning setting instead focuses on AutoResearch's research-decision behavior under externally defined evaluation criteria.

\subsection{Can AutoResearch Turn a Research Idea into Measurable Progress?}

We first evaluate whether an idea produced by AutoResearch can be translated into a concrete method and yield measurable progress under controlled experimentation.
We consider bidirectional image--text retrieval on the Remote Sensing Image Captioning Dataset (RSICD)~\citep{lu2018exploring}, where performance is measured by mean Recall (mR) across both retrieval directions.
The research objective is to examine whether combining global image--text understanding with progressively finer local grounding can improve retrieval quality.

\begin{figure}[ht]
\centering
\includegraphics[width=\linewidth]{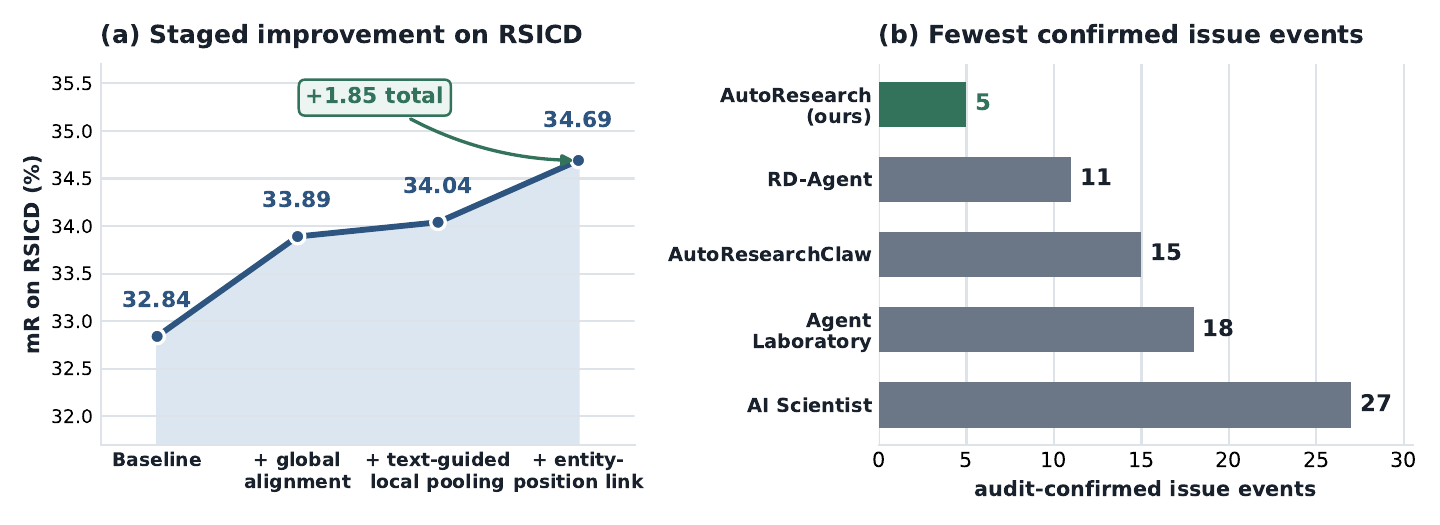}
\caption{
Evaluation on RSICD.
(a) The Idea Forge-generated method is introduced stage by stage under a fixed evaluation protocol, improving mR from $32.84$ to $34.69$ ($+1.85$).
(b) AutoResearch records $5$ audit-confirmed issue events, the fewest among the autonomous research systems compared.
}
\label{fig:retrieval}
\end{figure}

Idea Forge produces and cross-validates a staged hypothesis consisting of three components: stronger global image--text alignment, text-guided local feature aggregation, and explicit entity--position association.
Rather than evaluating the proposal only as a complete system, AutoResearch introduces each component sequentially under the same experimental protocol.
As shown in~\cref{fig:retrieval}(a), mR increases from $32.84$ for the baseline to $33.89$, $34.04$, and finally $34.69$, yielding a total improvement of $+1.85$ mR.
The monotonic progression is important because it provides direct evidence that the generated research idea can be decomposed into experimentally testable mechanisms whose contributions remain measurable as the method is developed.

We further audit the complete research process under the common protocol introduced above.
As shown in~\cref{fig:retrieval}(b), AutoResearch records $5$ confirmed issue events, compared with $11$ for R\&D-Agent, $15$ for AutoResearchClaw, $18$ for Agent Laboratory, and $27$ for The AI Scientist.
The comparison indicates that the observed improvement is accompanied by a comparatively reliable execution trace, rather than being supported only by a final headline metric.
Together, the results show that AutoResearch can turn a generated research idea into incremental and independently observable experimental progress.

\subsection{Can AutoResearch Detect and Correct an Unreliable Result?}
\label{subsec:matmul}

We next evaluate whether AutoResearch can detect when an apparently successful experiment is not yet sufficiently supported by evidence.
We use a classical matrix-multiplication task with an explicit experimental contract: a $1024\times1024$ FP32 multiplication must complete within $200\,\mathrm{ms}$, remain within a relative error of $10^{-5}$ against two numerical references, and achieve a run-to-run coefficient of variation below $20\%$ over ten wall-clock trials.
Because both correctness and measurement stability are directly verifiable, this setting isolates whether the system accepts a favorable number too early or continues to interrogate the evidence.

\begin{figure}[ht]
\centering
\includegraphics[width=\linewidth]{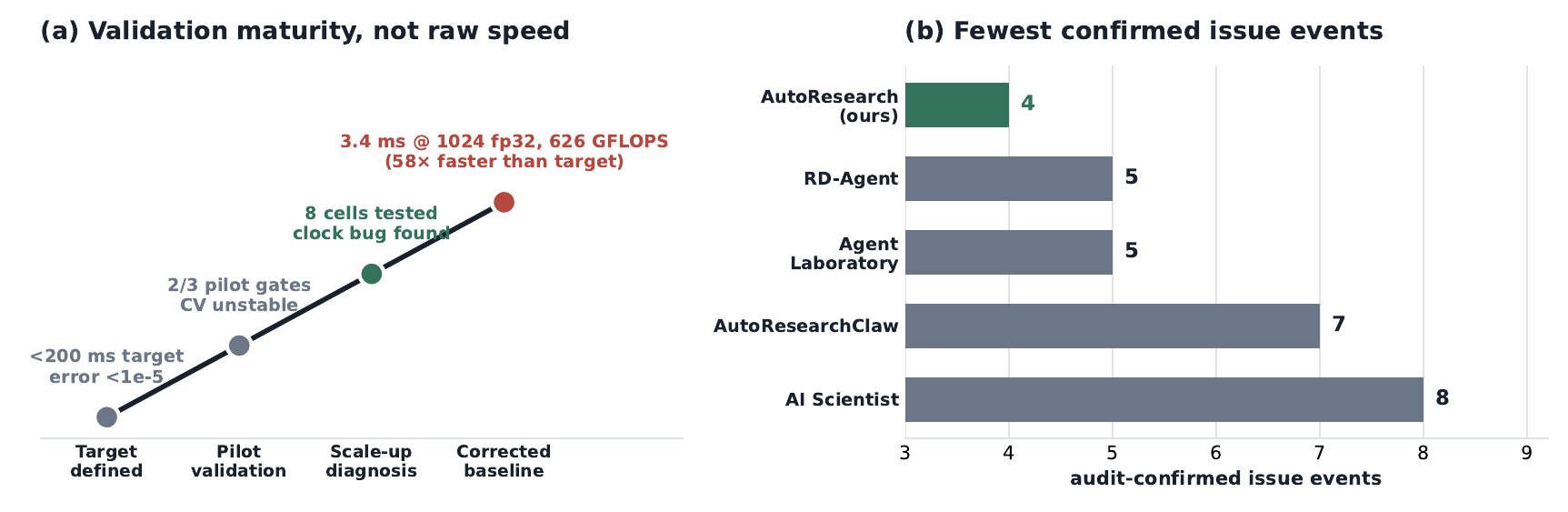}
\caption{
Validation of a $1024\times1024$ FP32 matrix-multiplication experiment.
(a) AutoResearch rejects an unstable pilot, diagnoses a timing error, and establishes a corrected $3.4\,\mathrm{ms}$ baseline after rerunning the experiment.
(b) AutoResearch records $4$ audit-confirmed issue events, the fewest among the five autonomous research systems compared.
}
\label{fig:matmul}
\end{figure}

As shown in Figure~\ref{fig:matmul}, AutoResearch initially satisfied the speed and numerical-accuracy criteria, but the pilot failed the stability requirement.
Rather than promoting the fast measurement into a research conclusion, the failed gate triggered a broader diagnostic sweep.
The subsequent analysis localized the inconsistency to a timing error: CPU time under multi-threaded BLAS was being conflated with elapsed wall-clock time.
After correcting the measurement procedure and rerunning the experiment, AutoResearch established a reproducible baseline of $3.4$,ms at $1024\times1024$ FP32, corresponding to $626$ GFLOPS and approximately $58\times$ the required speed margin.
The important outcome is therefore not the final runtime itself, but that an attractive intermediate result was rejected until its measurement process could be independently justified.

We further compare the complete execution traces under the same audit protocol.
AutoResearch records $4$ confirmed issue events, compared with $5$ for R\&D-Agent, $5$ for Agent Laboratory, $7$ for AutoResearchClaw, and $8$ for The AI Scientist.
Under this evaluation, AutoResearch produces the fewest audit-confirmed issues while also correcting the instability identified in its own pilot.
This case illustrates the intended role of evidence-grounded execution: experimental outputs are treated as provisional observations until they survive diagnosis and verification.

\subsection{Can AutoResearch Decide When to Continue, Revise, or Stop?}
\label{subsec:kaggle}

Finally, we examine whether AutoResearch can use experimental evidence to determine how a research direction should proceed, rather than treating continued experimentation as the default.
We consider three benchmark-driven machine-learning tasks: Titanic~\citep{cukierski2012titanic}, House Prices~\citep{decock2011ames,montoya2016houseprices}, and Disaster Tweets~\citep{howard2019disastertweets}.
For each task, AutoResearch starts from a low-cost baseline, iteratively evaluates concrete modifications, and compares the observed progress against a predefined target.

\begin{figure}[ht]
\centering
\includegraphics[width=\linewidth]{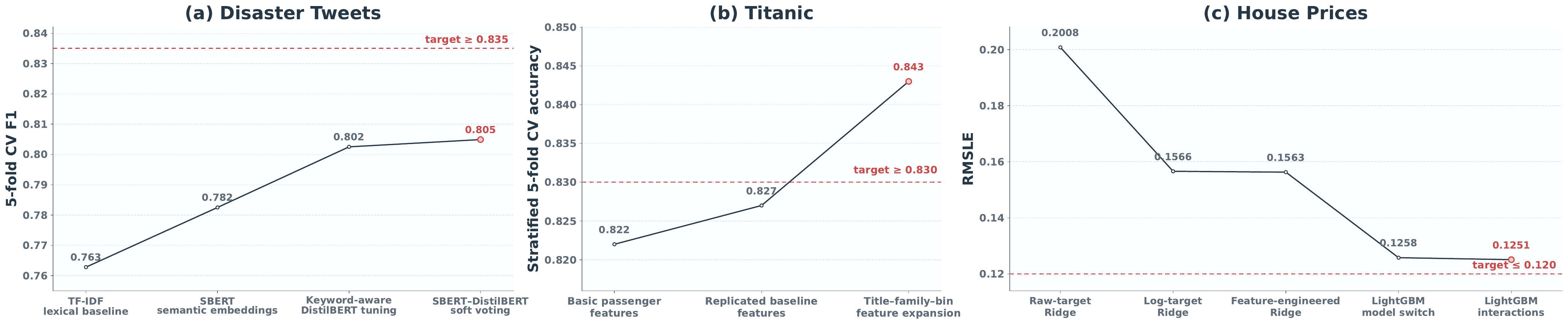}
\caption{
Experimental progress and evidence-conditioned decisions across three Kaggle tasks.
Titanic exceeds its target and supports scale-up, House Prices approaches its target and motivates further revision, while Disaster Tweets exhibits diminishing gains below its target and is terminated with the negative result retained.
}
\label{fig:kaggle_decisions}
\end{figure}

\cref{fig:kaggle_decisions} shows three distinct research trajectories.
On Titanic, feature expansion improves five-fold cross-validation accuracy from $0.822$ to $0.843$, exceeding the target of $0.830$ and supporting further scale-up.
On House Prices, successive changes reduce RMSLE from $0.2008$ to $0.1251$, approaching but not yet reaching the target of $0.120$; the remaining gap motivates continued revision rather than acceptance of the current solution.
In contrast, Disaster Tweets improves from $0.763$ to $0.805$ F1 but remains well below the target of $0.835$, with the final modifications yielding only marginal gains.
The system treats the direction as plateaued and terminates the run while retaining the negative result.

These trajectories illustrate that research closure in AutoResearch is conditioned on accumulated evidence rather than execution length.
A direction that reaches its objective can be expanded, one that remains promising can be revised, and one that ceases to make meaningful progress can be stopped without indefinitely searching for a positive result.
Together with the preceding evaluations, this result shows that AutoResearch uses experimental evidence not only to establish research conclusions, but also to determine what research action should follow.

\section{Discussion and Future Work}
\label{sec:discussion}

\textbf{AutoResearch is designed to optimize for justified research decisions rather than positive findings alone.}
Across the evaluated scenarios, progress may lead to scale-up, revision, falsification, or termination, as long as the resulting decision is supported by experimental evidence.
Negative and inconclusive outcomes are therefore treated as valid research results rather than failures of autonomous execution.

\textbf{The system can scale research exploration while retaining a selective generation--verification process.}
On a single server with dual Intel Xeon Platinum 8563C CPUs (98 cores / 196 threads), $8\times$ NVIDIA L20 GPUs, and $944\,\mathrm{GiB}$ memory, AutoResearch generated approximately 2584 candidate ideas within one week ($7 \times 24$ hours) of continuous operation.
After multi-model review and filtering, about 355 ideas entered the experimental queue, leading to roughly 22 automatically executed experiments and approximately 14 empirically validated ideas.
This operating point illustrates how broad idea exploration can be combined with progressively stricter selection and experimental verification.

\textbf{The next step is to turn individual research runs into a continuously evolving research process.}
The current system still depends on the coverage of external signals, the quality of accumulated domain knowledge, and the availability of explicit experimental criteria for verification.
A natural extension is to feed the verified evidence and research outcomes from each research cycle back into the knowledge state.
Specifically, let $\mathcal{K}_{t}$ denote the accumulated research knowledge at cycle $t$, $\mathcal{E}_{t}$ the verified experimental evidence produced during that cycle, and $\mathcal{Y}_{t}$ the resulting research outcomes, including supported, falsified, or inconclusive hypotheses.
The knowledge state can then be updated as
\begin{equation}
\mathcal{K}_{t+1}
=
U\!\left(
\mathcal{K}_{t},
\mathcal{E}_{t},
\mathcal{Y}_{t}
\right),
\end{equation}
where $U(\cdot)$ denotes the knowledge-update process.
This feedback allows subsequent research to benefit from both the evolving external research landscape and AutoResearch's own validated experience.
Combined with more adaptive multi-agent and swarm-style coordination, it provides a path toward a continuously improving research system while preserving the principles of \textit{Insight In} and \textit{Hallucination Out}.

\section{Conclusion}
\label{sec:conclusion}

AutoResearch formulates autonomous research as a two-stage process of \textit{Idea Generation} and \textit{Idea Execution}.
It grounds research ideas in evolving signals and domain knowledge, and grounds research conclusions in experimental evidence obtained through iterative execution and independent verification.
Across three representative research scenarios, AutoResearch demonstrates measurable idea development, correction of unreliable experimental results, and evidence-conditioned research decisions.
These results suggest that autonomous research should be evaluated not only by how much of the workflow is automated, but by whether the resulting research process remains grounded from idea formation to experimental conclusion.
AutoResearch operationalizes this principle as \textit{Insight In, Hallucination Out}.

\bibliographystyle{unsrtnat}
\bibliography{references}

\end{document}